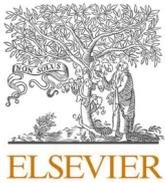
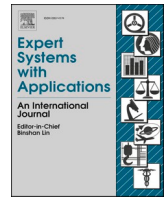

# SWP-LeafNET: A novel multistage approach for plant leaf identification based on deep CNN

Ali Beikmohammadi [a,*,1], Karim Faez [b,2], Ali Motallebi [b,3]

[a] *Department of Computer and Systems Sciences, Stockholm University, Stockholm, Sweden*
[b] *Department of Electrical Engineering, Amirkabir University of Technology (Tehran Polytechnic), Tehran, Iran*

ABSTRACT

Modern scientific and technological advances allow botanists to use computer vision-based approaches for plant identification tasks. These approaches have their own challenges. Leaf classification is a computer-vision task performed for the automated identification of plant species, a serious challenge due to variations in leaf morphology, including its size, texture, shape, and venation. Researchers have recently become more inclined toward deep learning-based methods rather than conventional feature-based methods due to the popularity and successful implementation of deep learning methods in image analysis, object recognition, and speech recognition.

In this paper, to have an interpretable and reliable system, a botanist's behavior is modeled in leaf identification by proposing a highly-efficient method of maximum behavioral resemblance developed through three deep learning-based models. Different layers of the three models are visualized to ensure that the botanist's behavior is modeled accurately. The first and second models are designed from scratch. Regarding the third model, the pre-trained architecture MobileNetV2 is employed along with the transfer-learning technique. The proposed method is evaluated on two well-known datasets: Flavia and MalayaKew. According to a comparative analysis, the suggested approach is more accurate than hand-crafted feature extraction methods and other deep learning techniques in terms of 99.67% and 99.81% accuracy. Unlike conventional techniques that have their own specific complexities and depend on datasets, the proposed method requires no hand-crafted feature extraction. Also, it increases accuracy as compared with other deep learning techniques. Moreover, SWP-LeafNET is distributable and considerably faster than other methods because of using shallower models with fewer parameters asynchronously.

## 1. Introduction

In agronomy, an important task is to identify and classify plants (with approximately 50,000 species) for botanical research and agricultural products (Hall et al., 2015; Kalyoncu & Toygar, 2015; Kumar et al., 2012) because plants serve as the backbone of all forms of life on Earth and provide humans with food and oxygen. As a result, it is necessary to have a proper perception of plants to help identify new or rare species of plants in order to improve the pharmaceutical industry, ecosystem balance, and agricultural productivity and sustainability (Cope et al., 2012).

In botany, plants are typically identified based on the shapes of their leaves and flowers. Botanists use the changes of leaf characters as a comparative tool to study the plants (Clarke et al., 2006; Cope et al., 2012) because the leaf characters of deciduous trees, annual plants, or year-round in evergreen perennials are available for observation and analysis. In fact, leaves are always employed to provide important detective characters to identify plants.

Analyzing plant leaves in specialized botanical laboratories, in addition to the need for expertise and knowledge, is a costly and time-consuming task. To solve this problem, many researchers of computer vision have used images of leaves as a tool to identify species of plants

* Corresponding author.
*E-mail addresses:* beikmohammadi@dsv.su.se (A. Beikmohammadi), kfaez@aut.ac.ir (K. Faez), ali3motallebi@aut.ac.ir (A. Motallebi).
[1] ORCID: 0000-0003-4884-4600.
[2] ORCID: 0000-0002-1159-4866.
[3] ORCID: 0000-0002-1845-4997.

(Hall et al., 2015; Kalyoncu & Toygar, 2015; Kumar et al., 2012). Also, given scientific and genetic advances, many hybrid plants cannot be identified by ordinary individuals and even experts (Ferentinos, 2018). Therefore, machine vision and machine learning techniques are used more often to solve this problem (Kamilaris & Prenafeta-Boldú, 2018; Lee et al., 2015). Shape (Mouine et al., 2012; Neto et al., 2006; Xiao et al., 2010), texture (Cope et al., 2010; Naresh & Nagendraswamy, 2016; Tang et al., 2015), and venation (Charters et al., 2014; Larese et al., 2014) are the characters used mainly to identify the leaves of different species.

In computer vision, despite many attempts (e.g., the use of complicated compute vision algorithms), plant identification is still known as a challenging and unsolved problem because plants are characterized by shapes and colors, resembling others in nature. Moreover, the history of plant identification methods indicates that existing solutions depend significantly on the ability of experts to encrypt the domain knowledge. Regarding many of the morphological features predefined by botanists, researchers employ hand-engineering methods to define their dedicated characters. They seek specific methods and algorithms to extract a great deal of information for predictive modeling. Based on the efficiency of those methods and algorithms, a subset of very important features is selected to describe leaf data. Recently, deep learning methods have been employed for the automated extraction of features, something which has alleviated researchers' tasks in the explicit selection of features.

Deep learning includes a category of machine learning techniques consisting of numerous processing layers, making it possible to learn the representation of data concepts at multiple levels. Deep learning is an opportunity to create and extract new features of raw representation of input data without clearly defining what features are used and how they are extracted.

In recent years, deep learning has brought about significant development in machine learning and artificial intelligence. Since 2012, all the top ranks of ImageNet Large Scale Visual Recognition Challenge (ILSVRC), known as the machine vision world cup, have used deep neural networks. Moreover, all the top methods of MNIST handwritten digit image classification (21 errors in 10,000) and CIFAR natural images (less than five percent error) belong to deep neural networks.

As deep learning has been successful in various fields, it has recently entered agronomy and food production. The agricultural applications of deep learning include plant disease recognition, Earth coverage classification, product-type classification, plant identification, plant phenology recognition, root-soil separation, production estimation, fruit enumeration, obstacle recognition, weed recognition, product recognition and classification, soil moisture forecast, animal studies, and weather forecast (Kamilaris & Prenafeta-Boldú, 2018).

Regarding plant identification, various studies have focused on the methods and algorithms maximizing the use of leaf databases. This fact leads to a norm indicating that leaf features may change with different leaf data and feature extraction techniques. There has always been a vague subset of features representing leaf data.

**Contribution:** The main purpose of this paper is to provide a reliable framework and a suitable alternative to botanical laboratory methods for identifying plants. In particular, instead of addressing a display of features in the same way as previous approaches, this study first analyzed the steps taken by a botanist on a leaf for plant recognition. Then we model the same process. For this purpose, deep learning is employed to extract specific features by limiting the input. Specifically, we have designed a system consisting of three separate and distributable models based on convolutional neural networks (CNNs), where the first and second models are optimized as much as possible to be shallow and fast so that they can be implemented on portable devices such as mobile phones. We used deep transfer learning on a pre-trained model with fast and accurate performance for the third model. Although, at first glance, applying deep learning seems to prevent the system from being interpretable, by visualizing the various layers of the proposed system, we can give botanists the confidence that the proposed system works the way they do. We have also measured the accuracy of the proposed method on two well-known datasets in this field, which indicates our better performance than the state-of-the-art methods. Interpretability, high asymmetric speed due to the multi-stage execution, and implementable in portable devices are other advantages of the proposed method. Hence, this study includes all the realistic assumptions that originate from the practical usage and goal of such a system.

The remaining sections are organized as follows: Section 2 presents a review of the literature on plant leaf classification. Then, Section 3 introduces the proposed method inspired by a botanist's behavior in plant recognition from leaves. Section 4 includes performance analysis and reports the simulation results of the suggested approach. For this purpose, well-known datasets such as Flavia (Wu et al., 2007) and MalayaKew (MK) (Lee et al., 2017; Lee et al., 2015) are employed. In addition, this section also compares the proposed method with several methods proposed by other researchers, reviewed in Section 2. Section 5 presents the research conclusion and makes suggestions for future studies.

## 2. Related studies

The representation of leaf features is a critical component of leaf identification and classification algorithm. All the existing methods follow two general approaches to the representation of features extracted from images of leaves to classify species: hand-crafted feature extraction (Charters et al., 2014; Naresh & Nagendraswamy, 2016; Neto et al., 2006) and deep learning feature extraction (Grinblat et al., 2016; Lee et al., 2017; Lee et al., 2015; Liu et al., 2015).

In practice, designing hand-crafted features depends significantly on the ability of computer vision experts to encrypt the morphological features predefined by botanists (Lee et al., 2017). Nevertheless, deep learning features are capable of automated learning, benefiting from the advantages of deep learning algorithms. Hence, deep learning methods for leaf identification have recently become more popular. Thus, this work reviews related deep learning-based studies.

There have been considerable developments in deep learning methods. The deep learning method introduced by (Su et al., 2014) is employed to reduce data dimensions in CNNs (Chollet, 2017) and deep belief networks (Simonyan & Zisserman, 2014). These methods have extensively been used for image classification, speech recognition, and object recognition (Angelov & Sperduti, 2016). Learning-based representation and particularly deep learning have introduced the concept of end-to-end learning by employing trainable feature extractors and trainable classifiers (Szegedy et al., 2017; Szegedy et al., 2016).

Recently, a few feature extraction methods for leaf classification have been proposed through deep learning (Barré et al., 2017; Bodhwani et al., 2019; Grinblat et al., 2016; Hedjazi et al., 2017; Hu et al., 2018; Lee et al., 2017; Lee et al., 2015; Liu et al., 2015; Sun et al., 2017). (Liu et al., 2015) used a conventional CNN for feature extraction, then they employed a support vector machine (SVM) to classify images of leaves.

(Grinblat et al., 2016) first, segmented the venation pattern of a leaf through a hit-or-miss transform (UHMT) to obtain the segmented binary venation images. Then they trained a CNN with these segmented binary images instead of the main input images. In fact, deep learning and hand-crafted feature extraction are integrated into this method. (Barré et al., 2017) developed the CNN-based LeafNet network and evaluated it on Flavia, LeafSnap, and Foliage resulting in 97.9%, 86.3%, and 95.8% accuracy rates, respectively.

(Lee et al., 2015) proposed the DeepPlant network for the recognition of images of plant leaves. They also used a deconvolutional network (DN) (Shelhamer et al., 2016) to obtain an insight into the designated features resulting from the CNN model. According to them, deep learning should be used in either a bottom-up or top-down method for plant identification. First, they proposed a CNN model for the automated learning of feature representation for plant species, a method that meets the need to design hand-crafted feature extraction in previous methods. Then they identified and recognized the representation of features obtained by the CNN model through a DN-based visualization strategy. This means avoiding the CNN model as a black box solution. It also gives researchers an insight into how an algorithm "observes" or "perceives" a leaf. Finally, they collected a new dataset named MalayaKew (MK) through complete labeling and made it available to other researchers.

**Table 1**
An Overview of Related Works on Feature Extraction through Deep Learning in Leaf Classification for Plant Identification.

| Publications | Method |
| --- | --- |
| (Liu et al., 2015) | CNN & SVM |
| (Lee et al., 2015) | DeepPlant (CNN, DN) |
| (Grinblat et al., 2016) | UHMT & CNN |
| (Hedjazi et al., 2017) | Pre-trained AlexNet |
| (Sun et al., 2017) | ResNet26 |
| (Barré et al., 2017) | LeafNet |
| (Lee et al., 2017) | TwoCNN (CNN, DN) |
| (Hu et al., 2018) | MSF-CNN |
| (Bodhwani et al., 2019) | ResNet50 |

Empirically, the method proposed by (Lee et al., 2015) outperformed the state-of-the-art solutions based on hand-crafted feature extraction in classifying 44 different plant species. Therefore, it indicates that feature learning through CNN can provide a better representation of images of leaves than hand-crafted feature extraction. In addition, tests are conducted on the images of entire leaves and images of leaf segments, the results of which show that venation is an important feature in the identification of different plant species and outperforms conventional solutions.

In the paper reviewed by (Lee et al., 2017), perfected their previous study significantly and proposed a two-stream convolutional neural network (TwoCNN), including two feature learning streams trained on the entire and segments of images, respectively. Like the paper reviewed by (Lee et al., 2015), first defined a method for quantifying the necessary features to show leaf data and trained a CNN based on raw leaf data to learn a resistant representation of images of leaves. Then they employed a DN method to observe how to describe leaf information through CNN. They determined the characters of features of each CNN layer quantitatively and found out that the layer transfer network would reach from a general mode to specific types of leaf features. They employed feature visualization techniques to explore, analyze, and perceive the most important subset of features. Interestingly, this study follows the definitions of characters given by botanists to classify plant species.

According to (Lee et al., 2017), CNNs trained on both the entire leaves and leaf patches indicated different contextual information from leaf features. As a result, they classified the information as global features, describing the entire leaf structure and local features focusing on venation. Finally, they proposed a new hybrid model titled TwoCNN to extract global–local features for leaf data. This model merges the information through two trained CNNs by using different data formats extracted from the same species. Although the proposed global–local feature extraction hybrid models can enhance the distinctive information of plant classification systems at different scales (i.e., both the entire images and patches of images), the training process requires a more complicated sample set because both the entire images and patches of images must be prepared.

Most of the existing leaf identification methods normalize the entire images of a plant leaf at the same rate and identify them on the same scale, a fact that leads to inappropriate results. For this reason, (Hu et al., 2018) tried to integrate multi-scale features with CNNs and develop a model titled MSF-CNN to classify plant leaves on multiple scales. The idea of integrating multi-scale features with CNNs is proposed by (Du & Gao, 2017) and also (Rasti et al., 2017), introducing multi-scale convolutional neural networks (MSCNNs). These MSCNNs consist of multiple branches of learning features on different scales.

In the paper reviewed by (Du & Gao, 2017), each branch learns information from patches of images in various sizes. However, in the paper reviewed by (Rasti et al., 2017), each branch learns information from the input images in different sizes. The main difference between MSCNNs and MSF-CNN proposed by (Hu et al., 2018) is that the multi-scale features learned by MSCNNs are integrated into one layer. Nonetheless, they are integrated in a step-by-step manner in an MSF-CNN. Hence, MSF-CNN is cost-effective and economical because it needs no multiple branches of learning features.

In the processing steps of the method proposed by (Hu et al., 2018), an input image is first transformed into several low-resolution images by reducing the sampling rate through a list of two-way interpolation operations. Then these input images are given to the MSF-CNN architecture on different scales through various strides to learn distinctive features at diverse depths. In this step, the integration of features is performed through concatenation operation between two different scales. In fact, these operations interconnect the mappings of features learned from multi-scale images in a channel view. Along the depth of the MSF-CNN, multi-scale images are gradually moved, and corresponding features are merged. Eventually, the last MSF-CNN layer collects distinctive information to obtain the final features to predict the plant species of an input image. (Hu et al., 2018) conducted a few experiments on both famous datasets of this area, i.e., MalayaKew (MK) and Leafsnap, and claimed that the MSF-CNN outperformed the most advanced methods of leaf identification. They also performed a mutual dataset evaluation to represent the generalizability of their method by training MK and testing Leafsnap.

(Hedjazi et al., 2017) modified a trained model for plant recognition. They showed how to use a mode, previously trained on a large dataset, for a small dataset. They did not train their model from the beginning. Instead, they selected a CNN model previously trained on ImageNet. They worked on ImageClef2013, a dataset including images with clean backgrounds. Given the deficiency of training data, a fitness problem is possible. Thus, they used transfer learning to avoid this problem. They fine-tuned an AlexNet model with the help of the Caffe framework and obtained 71.17% of accuracy on validation datasets and 70.0% of accuracy on testing datasets.

(Sun et al., 2017) analyzed BJFU100 consisting of 100 species of ornamental plants, each of which has 100 different pictures of 4208x3120 pixels. They compared ResNet26 (with 26 layers) with the ResNet models of 18, 34, and 50 layers and introduced ResNet26 as the best solution for the optimization and capacity problems. According to them, ResNet26 has enough trainable parameters to learn distinctive features. Their model obtained 91.78% of accuracy.

(Bodhwani et al., 2019) analyzed Leafsnap with 185 different plant species and employed the residual deep learning framework with 50 layers (in five classes) to classify the dataset. Their proposed model obtained 93.09% of accuracy.

Table 1 depicts an overview of the literature review of papers using feature extraction with deep learning.

By reviewing the related studies, several salient points can be made. First, the challenge in this area that researchers deal with is the high resemblance of leaves of diverse plants regarding shape, color, and morphological variations such as changes in size, texture, shape, and venation. In addition, existing many dimensions of numerous species of plants or even the same species of one plant in various growth conditions or photography periods are considered as other challenges. Second, applying deep learning-based methods is the current trend to overcome these challenges, where researchers have succeeded in reaching acceptable accuracy in the identification of plant leaves. However, as a third point, in the current trend, the uninterpretability and unreliability of the system when observing new species have made them never a viable alternative to botanical laboratory methods. Specifically, botanists prefer to rely on a system that they are sure works like them, not a black box system.

Fourth, the current approaches have achieved better accuracy by making the model more complex and costly in terms of computation. However, since one of the main aims of developing such systems is using them on-site, attention should be paid to their applicability on portable devices, being efficient in terms of computational cost and memory usage in general. That is why, by introducing a multi-stage system while improving performance, we have paid attention to the interpretability of the results and the possibility of using it on-site. Hence, our proposed system can be a quick and efficient alternative to botanical laboratory methods.

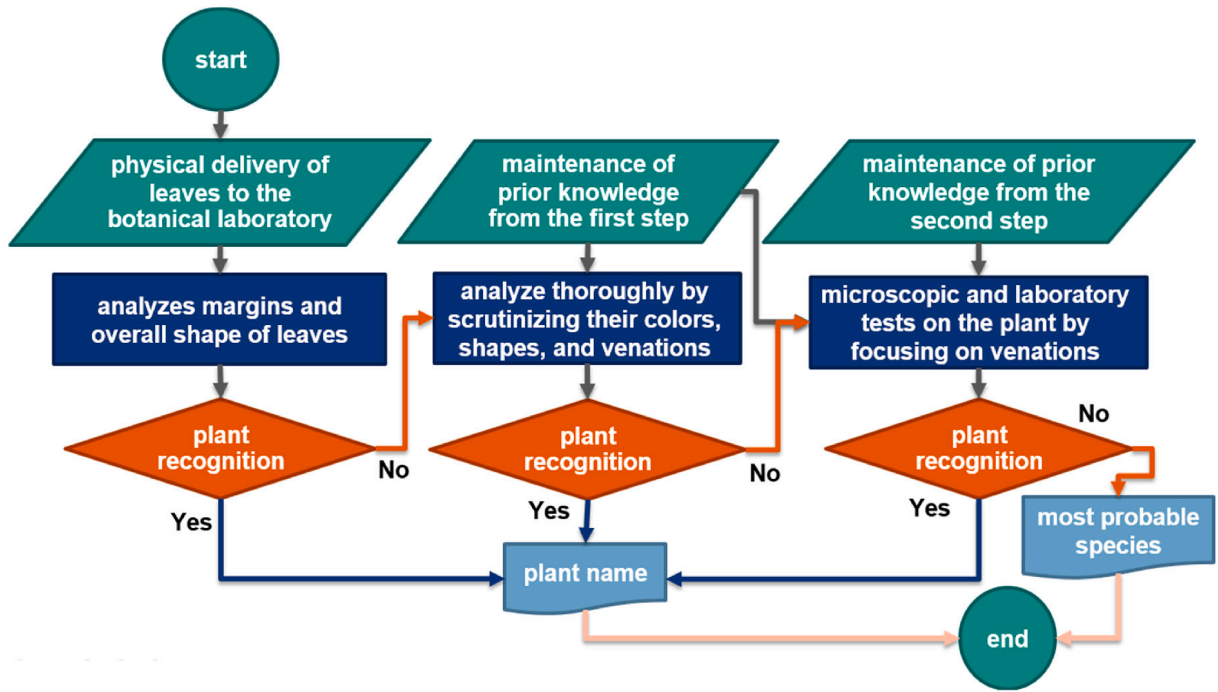

**Fig. 1.** Species Recognition Process Used by Botanists.

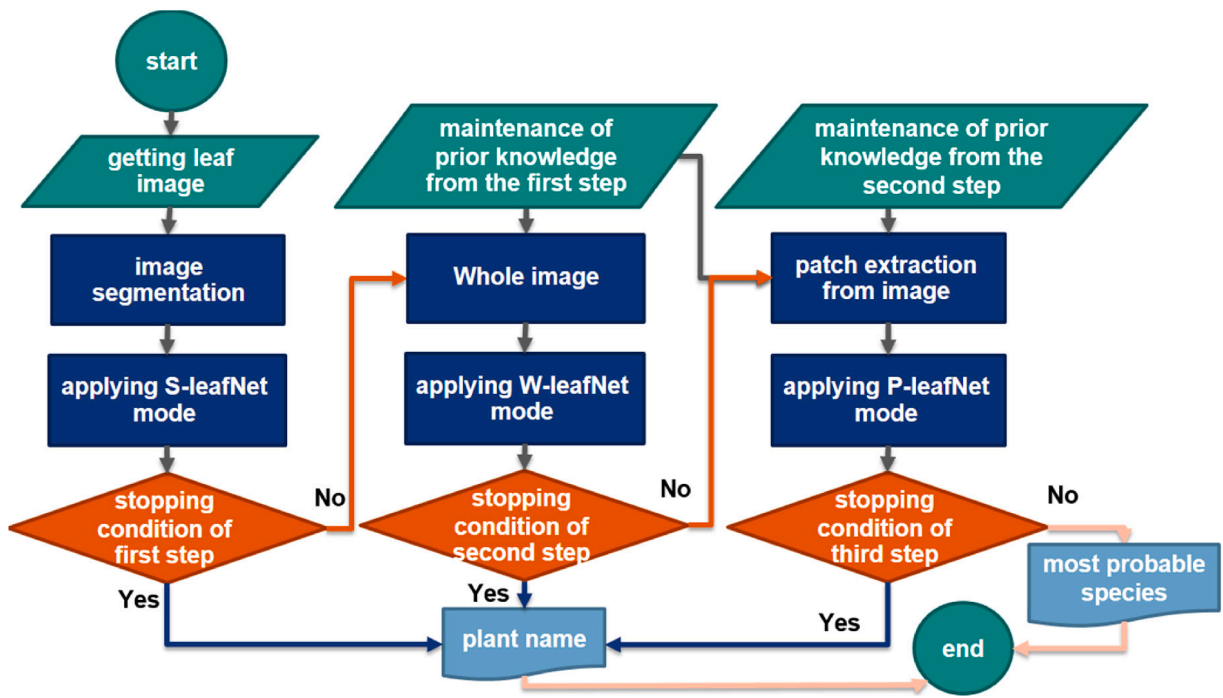

**Fig. 2.** Proposed Method for Species Recognition: SWP-LeafNET.

## 3. The proposed method

This section first addresses the process of recognizing a plant species by botanists. Then a method is proposed to show the maximum resemblance to a botanist's behavior. For this purpose, the proposed method is introduced in general and in detail by describing the structures of all three architectures thoroughly, in addition to the termination conditions at each step.

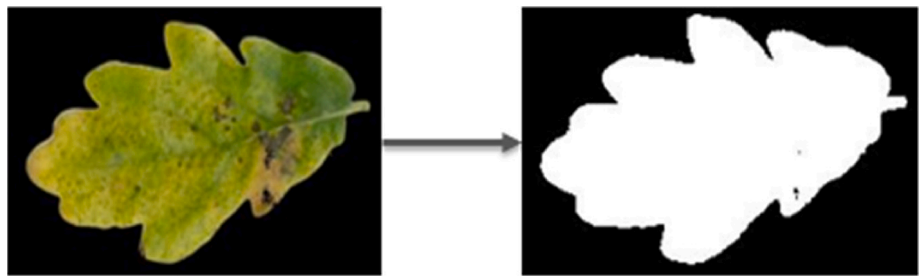

**Fig. 3.** Preprocessing Performed in the First Step of the Proposed Method.

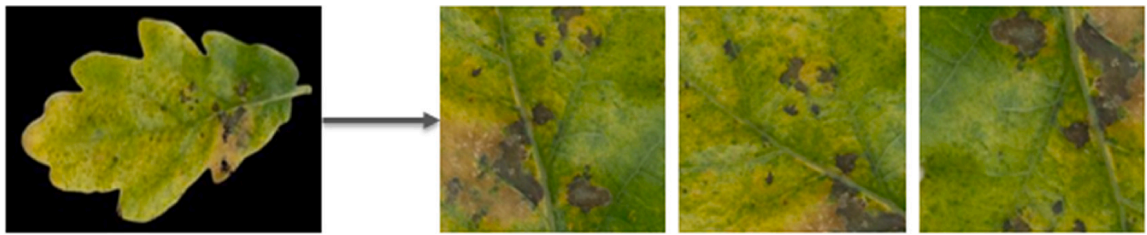

**Fig. 4.** Preprocessing Performed in the Third Step of the Proposed Method.

*3.1. Species recognition process proposed by botanists*

According to the flowchart shown in Fig. 1, botanists should take the following steps to recognize plant species:

First, some leaves should be cut and used as specimens, which should immediately be taken physically to a botanical laboratory. Then a botanist analyzes leaf appearance and general features, including its margins and overall form. If the initial analysis can help distinguish the leaf from other species of the geographical region, the botanist provides an inquirer with the plant species and information. If the initial analysis is prone to uncertainty, the botanist remembers the specimens and starts to analyze them thoroughly by scrutinizing their colors, shapes, and venations. If the botanist can turn the uncertainty of the first step into certainty by integrating the knowledge from the first step with the information of the second step, the plant species recognition process is completed. However, suppose the botanist is still skeptical about several plant species. In that case, he/she starts running microscopic and laboratory tests on the plant by concentrating on venations in order to provide the inquirer with the accurate name and information of the plant species by combining the knowledge previously obtained from the first and second steps.

The botanists may not determine the name and characters of the plant species even after the third step for various reasons, such as the low quality of the laboratory specimen and genetic changes applied to the plant. In this case, he/she provides the inquirer with a list of the most plausible plant species resembling the one delivered to the laboratory. This simple process is based on striking a balance between simplicity in recognizing specific species and accuracy in recognizing complicated species. This process can prevent wastage of time and resources for analyzing species, which can easily be distinguished from other species. Also, it facilitates the accurate recognition of suspicious species, which are hard to distinguish from others.

However, in general, the process of plant detection in the botanical laboratory requires cutting the plant leaves and transporting them to the laboratory, which is sometimes impossible due to the rarity of the plant species. In addition, this process is costly and time-consuming and requires a botanist compared to machine vision-based approaches. For this reason, we intend to develop an interpretable yet fast-paced, accurate, and distributable system inspired by botanists' approach for identifying plant leaves. Our method can be used on-site and is, therefore, cost and time-efficient. In addition, it does not require botanical expertise.

However, it can also help expert botanists as the first step in diagnosing.

*3.2. The proposed method for species Recognition: SWP-LeafNET*

After presenting a complete introduction to the species recognition process used by a botanist, it is decided to propose a method bearing the closest resemblance to increase the confidence of having an interpretable and reliable system. Fig. 2 illustrates a flowchart for a simple perception of the proposed method. This flowchart also eases comparing the suggested approach with the botanist's plant recognition process.

The following steps are recommended to recognize species through the proposed method. First, an image of the plant under study should be taken as a specimen, which should be given as input to the system. As shown in Fig. 3, the system then performs a preprocessing procedure only to analyze the general appearance of leaves, including margins and forms. In this preprocessing procedure, the leaf background is colored in black, whereas the entire leaf is colored in white. If the plant species can be distinguished from the other species of the same geographical region after applying the image to the first model known as S-LeafNET when the termination conditions of the first step are met, then the system provides the inquirer with the plant species and plant information. However, suppose the termination conditions of the first step are not met because the first model is skeptical about different species. In that case, the most plausible species are remembered because the final answer lies definitely among them. Then the proposed system starts to analyze colors, shapes, and venations of leaves more thoroughly than before.

For a more in-depth analysis of the plant, a colorful image of the leaf is given as an input to the second model known as W-LeafNET. The plant species recognition process ends if the uncertainty of the first step changes to certainty when the initial knowledge from the first step is integrated with the second step's information to meet the termination conditions of the second step. However, if the system is still skeptical about several plant species, it starts to run a microscopic analysis of the plant by focusing on venations in order to provide the inquirer with the accurate name and information of the plant species by combining the previous knowledge resulting from the first and second step and the outputs of the final step.

As shown in Fig. 4, for the microscopic analysis of the plant in the third step, a specific and adjustable number of patches are first extracted automatically from the original image. The resultant image should entirely be enclosed within the leaf. Then each of these patches is given as the input to the third model known as P-LeafNET. The plant species recognition process successfully ends if the combination of initial knowledge obtained from the first and second steps and the third step's information can meet the third step's termination conditions.

**Fig. 5.** The First Model Architecture: S-LeafNET.

**Fig. 6.** The Second Model Architecture: W-LeafNET.

**Fig. 7.** The Third Model Architecture: P-LeafNET.

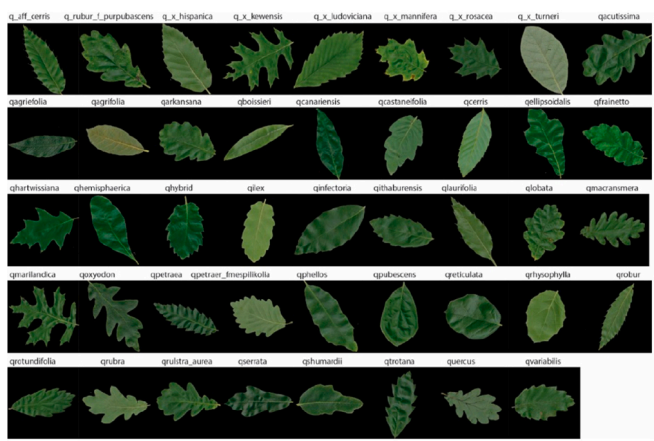

**Fig. 8.** Leaf Images with the Name of Each Plant Pertaining to MalayaKew (MK) (Lee et al., 2017; Lee et al., 2015).

Finally, suppose the system cannot determine the name and characters of a plant species even after the third step for different reasons such as the low-quality image of leaves or genetic changes applied to the plant species. In that case, a list of the most plausible species resembling the delivered species is given as output. Like what was discussed regarding the plant recognition process by a botanist, the proposed method can strike a balance between speed and accuracy. In other words, unlike all the state-of-the-art methods, we have an asymmetric execution in our system. This means that the easier it is to identify the plant species, the faster the decision will be made, just like the process by a botanist.

Although there is no theoretical rule in designing CNN-based models, during designing the S-LeafNET, W-LeafNET, and P-LeafNET models, we seek to optimize them in terms of memory usage, required computing, and storage resources by minimizing the numbers of computing layers while satisfying a good performance as much as possible. Based on this strategy, more than 100 different models have been designed and tested, of which the three models mentioned below are the best.

### 3.3. The first model, termination conditions, and initial knowledge transferred to the next steps

Since the input of the first model is a binary image, it is possible to design a simple, shallow, CNN-based model that can be used in portable devices with limited computational and storage resources. According to Fig. 5, the S-LeafNET model consists of five CBR layers (convolutional, Batch normalization (Ioffe & Szegedy, 2015), and ReLU activation function layers) as well as five pooling layers. Except for the last pooling layer, which is an average pooling layer, all of them are of max pooling. The purpose of the pooling layers is to reduce the dimensions of the hidden layer to reduce computational costs and increase the execution speed. Although the max pooling layer retains sharper features, the reason for using the average pooling layer as the last pooling layer goes back to our need to have global features rather than sharp features in the S-LeafNET model. This layer, in fact, like a low-pass filter, leads our model to pay more attention to general features. In addition, the dropout layers with different rates, L2 regularization, and batch normalization are included to help prevent the overfitting problem. In this way, we are more confident of the model's generalizability and the absence of bias to the dataset.

To be more specific, in the model input, a picture is first converted into a binary image through preprocessing. In this binary image, the leaf is colored in white to be distinguished from the black background. The input picture size is 128x128 pixels, and its binary version has a depth of 1. The first CBR layer uses 64 convolutional filters, sized 3x3, in a strider of 1 and input dimensions retaining padding in the output. As a result of applying this layer to the input, the output will be sized 128x128 in a depth of 64. After passing through the convolutional layer, batch normalization and ReLU activation function are applied to the output dimensions without changes in sizes.

In the second layer, max pooling is applied to the first layer with 2x2 dimensions and a strider of 1, as a result of which dimensions will decrease to 64x64 with a depth of 64. The third layer uses the same parameters as the first CBR layer but with a different number of filters, i.e., 128. As a result, the output dimensions will be 64x64, with a depth of 128. In the fourth layer, max pooling is applied to the third layer with 2x2 dimensions and a strider of 1, as a result of which dimensions will decrease to 32x32 with a depth of 128. Then the dropout layer (Srivastava et al., 2014) is used with a dropout rate of 0.1. The fifth layer uses the same parameters as the first CBR layer but with a different number of filters, i.e., 160. As a result, the output dimensions will be 32x32, with a depth of 160.

The sixth layer employs max pooling with the same parameters as the fourth layer, and the dimensions will decrease to 16x16 with a depth of 160. Then the dropout layer is used with a rate of 0.2. The seventh layer uses the same parameters as the first CBR layer with a different number of filters, i.e., 224. As a result, the output dimensions will be 16x16, with a depth of 224. The eighth layer uses max pooling with the same parameters as the fourth layer and decreases the dimensions to 8x8 with a depth of 224. Then the dropout layer is used at a dropout rate of 0.3. The ninth layer utilizes the same parameters as the first CBR layer but with different numbers of filters, i.e., 256. As a result, the output dimensions will be 8x8, with a size of 256.

Finally, the average pooling of 2x2 dimensions is applied to the ninth layer with a strider of 1 in the tenth layer. As a result, sizes will decrease to 4x4 with a depth of 256. Then a dropout layer is used at a rate of 0.4. After that, a multi-class sigmoid (also known as Softmax), including the same number of classes used in datasets, is employed to finish designing the S-LeafNET model. It should be mentioned that the total number of parameters in this model is approximately 1.3 million. Except for 1664 parameters, the rest of the parameters will be trained. Moreover, the model weights require 4.9 megabytes of storage space. Hence, this model can be implemented in portable devices.

About termination conditions, one of the bellow conditions should at least be met so that the system can decide to stop the process of analysis, validate the result of the first model, and introduce it as the final result. These conditions are as follows:

**Table 2**
Different Values of Parameters for Termination Conditions and Previous Knowledge Transfer for MalayaKew (MK) and Flavia.

| Parameter | MK Value | Flavia Value | Comments |
| --- | --- | --- | --- |
| min_prob_seg | 0.98 | 0.95 | Minimum probability for the class predicted by the first model |
| min_delta_seg | 0.95 | 0.91 | Minimum probabilistic distance of the predicted class from the second plausible class predicted by the first model |
| top_seg | 10 | 6 | The first N presumption of the first model on the right class |
| min_mean_L1L2pred | 0.80 | 0.70 | Minimum mean of both the first and second models in case of the sameness of the classes predicted by them |
| min_prob_whole | 0.89 | 0.78 | Minimum probability for the class predicted by the second model |
| min_delta_whole | 0.85 | 0.60 | Minimum probabilistic distance of the predicted class from the second plausible class predicted by the second model |
| top_whole | 6 | 10 | The first M presumption of the second model on the right class |
| P | 7 | 7 | The number of image patches generated for the third model |
| min_mean_L1L3pred | 0.60 | 0.60 | Minimum mean of both the first and third models in case of the sameness of classes predicted by the first and third models |
| min_mean_L2L3pred | 0.60 | 0.60 | Minimum mean of both the second and third models in case of the sameness of classes predicted by the second and third models |
| min_prob_ patch | 0.95 | 0.95 | Minimum probability for the class predicted by the second model |
| min_delta_ patch | 0.85 | 0.85 | Minimum probabilistic distance of the predicted class from the second plausible class predicted by the third model |
| vote_rate | 0.71 | 0.71 | Minimum rate after initial voting between the classes predicted by the third model on P image patches |
| vote_merge_rate | 0.56 | 0.56 | Minimum rate after general voting on the outputs of three models in case of insufficient certainty of initial voting and having the minimum sufficient probability through the first and second models |

- min_prob_seg: obtaining the least probability by the predicted class
- min_delta_seg: reaching the shortest probabilistic distance between the predicted class and the second possible class (the probabilistic distance between top-1 and top-2)

If one of these two conditions is met, the result of the S-LeafNET model will be presented as the final result. Otherwise, the system retains the initial knowledge obtained from the S-LeafNET model for the next models and starts using the W-LeafNET model. The initial knowledge obtained from the first step is described as follows:

- top_seg: transferring N first guesses from the first model on the correct class (storing predicted classes with the probability of each up to top-N)

*3.4. The second model, termination conditions, and the initial knowledge transferred to the next steps*

By following the strategy mentioned in Section 3.2 and testing many configurations, the W-LeafNET model, as shown in Fig. 6, consists of seven CBR layers with seven pooling layers, all of which of max pooling. In addition, different dropout rates have been used after the pooling layer from the second layer to one before the pooling layer to help prevent overfitting. In addition, L2 regularization and batch normalization are applied due to the same cause. Increasing the number of layers compared to the previous model is inevitable due to having the colored image as input and consequently more complexity and richness of the input. Also, as we seek to distinguish classes from detailed features by W-LeafNET, we have preferred using max pooling instead of average pooling, which highlights the most present feature.

W-LeafNET model input includes the colored image provided without preprocessing in the dimensions of 196x196 pixels. This image has a depth of 3 because of its RGB format. The first CBR layer consists of 64 convolutional filters sized 3x3 with a strider of 1. The padding maintains the input dimensions in the output. As a result of applying this layer to the input, there will be an output sized 196x196 with a depth of 64. After passing through the convolutional layer, batch normalization and ReLU activation functions are applied to the output dimensions of this layer without any changes. The second layer consists of max pooling sized 2x2 with a strider of 1 applied to the first layer. As a result, the dimensions will decrease to 98x98 with a depth of 64.

The third, fifth, seventh, ninth, and eleventh layers use the same parameters as the first CBR layer with a different number of filters, including 128, 160, 192, 224, and 320 filters, respectively. At the same time, the fourth, sixth, eighth, tenth, and twelfth layers use max pooling sized 2x2 with a strider of 1 with the dimensions of 49x49, 24x24, 12x12, 6x6, and 3x3, respectively. After applying max pooling to these layers, the dropout layer is used with different dropout rates of 0.1, 0.2, 0.3, 0.4, and 0.5, respectively. After passing through these layers, there will be 320 feature maps with dimensions of 3x3. The thirteenth layer benefits from the same parameters as the first CBR layer with a different number of filters, i.e., 256. As a result, the output dimensions will be 3x3, with a depth of 256.

Finally, the fourteenth layer uses the same max pooling parameters as the second layer. As a result, the dimensions will decrease to 1x1 with a depth of 256. After this layer, a multi-class sigmoid (also known as the Softmax) is used in the number of classes existing in datasets to finish designing the W-LeafNET model. It should be mentioned that there are approximately 2.3 million parameters in this model. Except for 2688 parameters, the others are trainable. The model weights require 8.9 megabytes of storage space. Therefore, although the designed model is slightly larger than the S-LeafNET model and has more parameters, it is still applicable on portable devices. Another point is that we never use both models simultaneously. In other words, first, only the S-LeafNET model is called and executed (which, in most cases, it is the only model we need; see the results in Section 4). If the results are not satisfactory, it is time to call the W-LeafNET model and check its output (which rarely happens).

**Table 3**
Accuracy Rates of Different Methods on MalayaKew (MK).

| S. No. | Publications | Method | Accuracy on the entire data (Accuracy on the processed data) |
| --- | --- | --- | --- |
| 1 | **Proposed method (Combine)** | **SWP-LeafNET** | **99.81%** |
| 2 | Proposed method (First model) | S-LeafNET | 96.59% (99.77%) |
| 3 | Proposed method (Second model) | W-LeafNET | 97.35% (100%) |
| 4 | Proposed method (Third model) | P-LeafNET | 95.35% (100%) |
| 5 | (Hu et al., 2018) | MSF-CNN | 99.05% |
| 6 | (Lee et al., 2017) | DeepPlant + SVM (linear) | 98.10% |
| 7 | (Lee et al., 2017) | DeepPlant + MLP | 97.70% |
| 8 | (Hall et al., 2015) | Combine (SVM (linear)) | 95.10% |
| 9 | (Hall et al., 2015) | HCF (SVM (RBF)) | 71.60% |
| 10 | (Hall et al., 2015) | HCF-ScaleRobust (SVM (RBF)) | 66.50% |
| 11 | (Kumar et al., 2012) | LeafSnap (NN) | 58.90% |
| 12 | (Kumar et al., 2012) | LeafSnap (SVM (RBF)) | 42.00% |
| 13 | (Yang et al., 2009) | SIFT (SVM (linear)) | 58.80% |

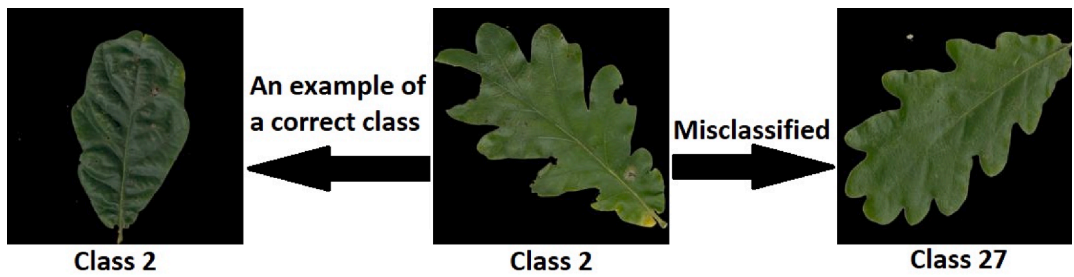

**Fig. 9.** A Sample of MalayaKew (MK) on Which Wrong Predictions are Made: (Left) A Sample of Class 2 Images; (Middle) A Sample of Class 2, Wrongly Predicted as a Part of Class 27; (Right) A Sample of Images from Class 27.

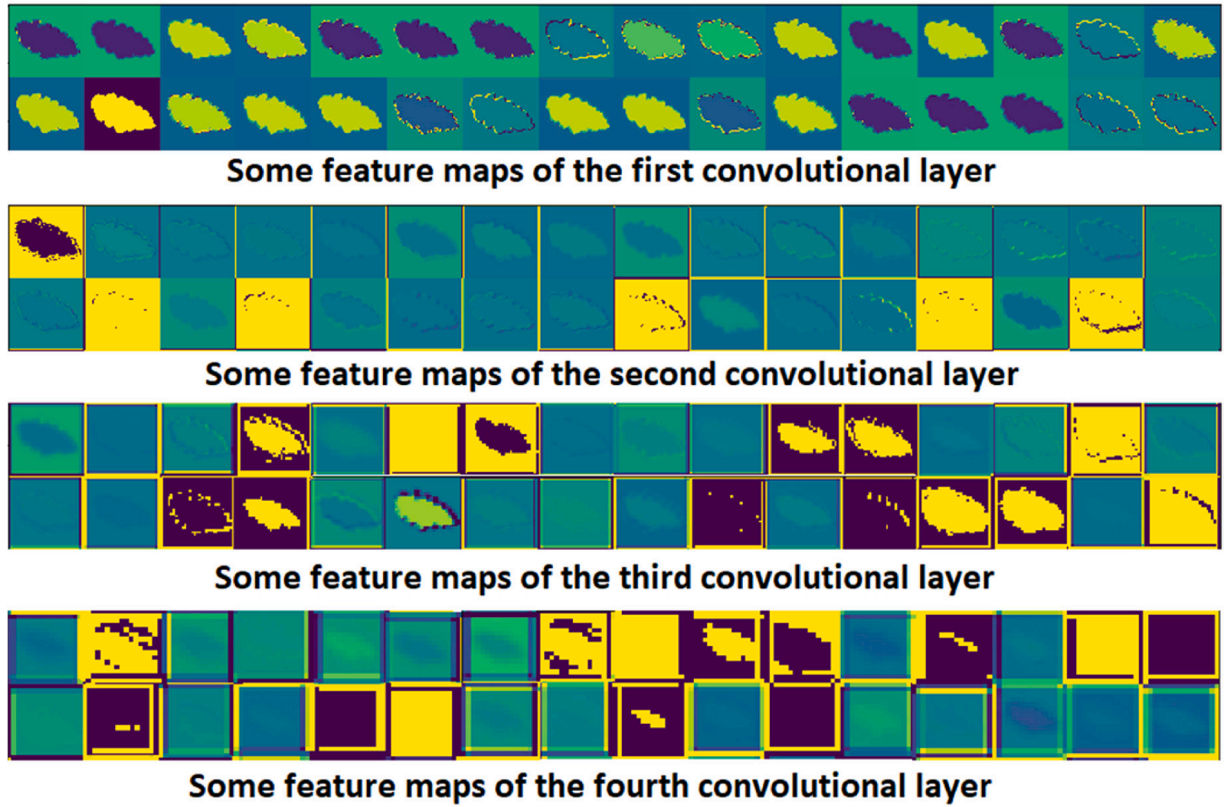

**Fig. 10.** A Number of Feature Maps of the First Four Convolutional Layers of the S-LeafNET Model Trained on the Black-and-White Images Obtained from MalayaKew (MK). The First Model Focused Only on the General Shape of the Leaf.

To stop running analyses and introduce the W-LeafNET model results as the final result, in addition to confirming the initial knowledge of the S-LeafNET model, the following termination conditions must be met:

- Availability of a class predicted by the second model of the first top-N model, and:
  o min_mean_L1L2pred: the sameness of the class predicted by the first model as the second model and obtaining the minimum mean by both the first and second models, or
  o min_prob_whole: obtaining the minimum probability by the predicted class, or
  o min_delta_whole: having the minimum probabilistic distance between the predicted class and the second possible class (probabilistic distance between top-1 and top-2)

If one of these conditions is met in addition to the first condition, the result of the W-LeafNET model will be regarded as the final result. Otherwise, the system starts using the P-LeafNET model by retaining the knowledge obtained from the first and second steps for the final step.

The initial knowledge obtained from the second step includes the following:

- top_whole: transferring the first M presumption from the second model on the accurate class (storing predicted classes along with the probability of each up to top-M)

According to the first and second termination conditions, using the initial knowledge resulting from the first step has a noticeable impact on the decisions made about the announcement of results obtained from the W-LeafNET model. The first condition, which should necessarily be true, indicates that although the S-LeafNET model failed to introduce the right class as the first choice, its first N presumptions are important because they affect the final result. The second condition is true when the S-LeafNET model fails to meet the least certainty conditions, despite presuming the right class. In this case, this presumption is confirmed by the W-LeafNET model to access the result less sensitively.

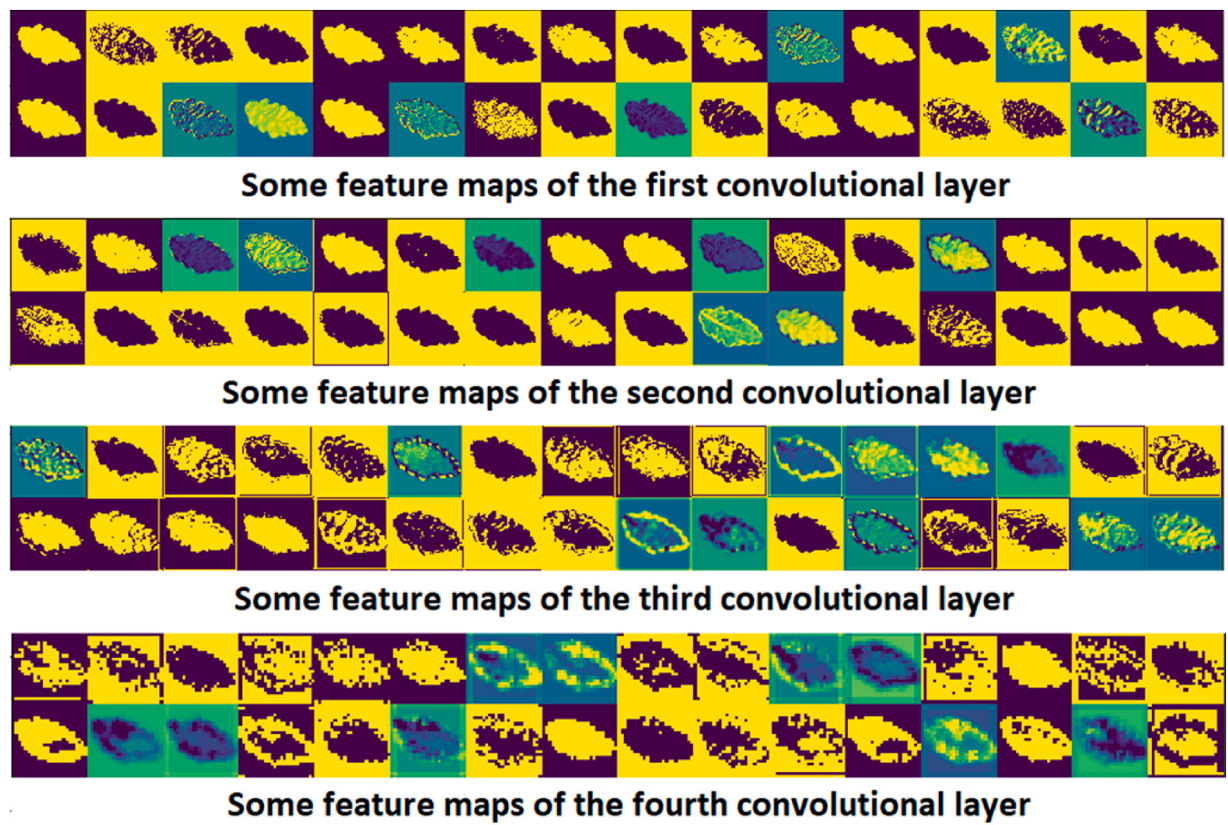

**Fig. 11.** A Number of Feature Maps of the First Four Convolutional Layers of the W-LeafNET Model Trained on the Colored Images Obtained from MalayaKew (MK). The Second Model Took an Extensive Look at the Leaf. However, It Focused on the Colors and Venations of Leaves, Although the General Shape is Much More Important to it.

*3.5. The third model, termination conditions, and introduction to the most plausible species*

Since we are dealing with patches of the image in this model, the distinctive features of which are very difficult to determine, we inevitably need a more complex CNN model. On the other hand, having a complex model requires more data for training, which is not available in botany. As a result, according to Fig. 7, the P-LeafNET model is developed based on deep transfer learning. This concept is known through two general models (Beikmohammadi & Faez, 2018):

- Retaining the major pre-trained network and updating weights based on the new training dataset
- Employing a pre-trained neural network for feature extraction and performing representation through a global classifier such as support vector machines

In this work, the first approach is employed to update the model weights to improve the P-LeafNET model's performance. Among the models winning the ILSVRC, MobileNetV2 (Sandler et al., 2018) is selected as the pre-trained network to develop the P-LeafNET model on the ImageNet dataset. The reason for this choice is due to the fact that it is completely in line with our model selection strategy. Specifically, it is a fast, relatively shallow, and compact model compared to other pre-trained models. Also, as its name implies, it can be used in portable devices. However, since this model will be rarely used (see the results in Section 4), it could be implemented on a server. Then we can access it by connecting to the server, sending image patches, and receiving the result.

Note that since the pre-trained model is to classify 1000 classes, which does not match the number of classes in our datasets, the last MobileNetV2 layer is dropped out, and a Softmax layer is replaced in proportion to the number of classes in the dataset. Then the entire new network is fine-tuned on the leaf dataset. As for the model input, a specific number of patches of a colored image is provided first. The patches are included merely inside the leaf. Then these patches are separately fed to the P-LeafNET model in the dimensions of 96x96. The input depth is three because image patches are of the RGB format. After passing through 88 different layers of MobileNetV2, a multi-class sigmoid (also known as Softmax) is used in the number of classes existing in datasets to finalize the third model. This model includes nearly 2.8 million parameters totally. Except for 34112 parameters, the others are trainable. The model weights require 10.9 megabytes of storage space.

The termination conditions of the third step should be met so that the system can decide to introduce the results of the P-LeafNET model as the final results, in addition to validating the initial knowledge resulting from the first and second models. After extracting P patches of the main image in this step, the mean of probabilities of classes on P image patches will be determined. Therefore, the P-LeafNET model output is a probability vector, including the probability of each class, for P image patches. The termination conditions are as follows:

- The presence of a class predicted by the third model except for the first top-N model and the second top-N model, and
  o min_mean_L1L3pred: the sameness of the class predicted by the first model as the one predicted by the third model and obtaining the minimum mean through both the first and third models, or
  o min_meanL2L3pred: the sameness of the class predicted by the second model as the one predicted by the third model and obtaining the minimum mean through both the second and third models, or

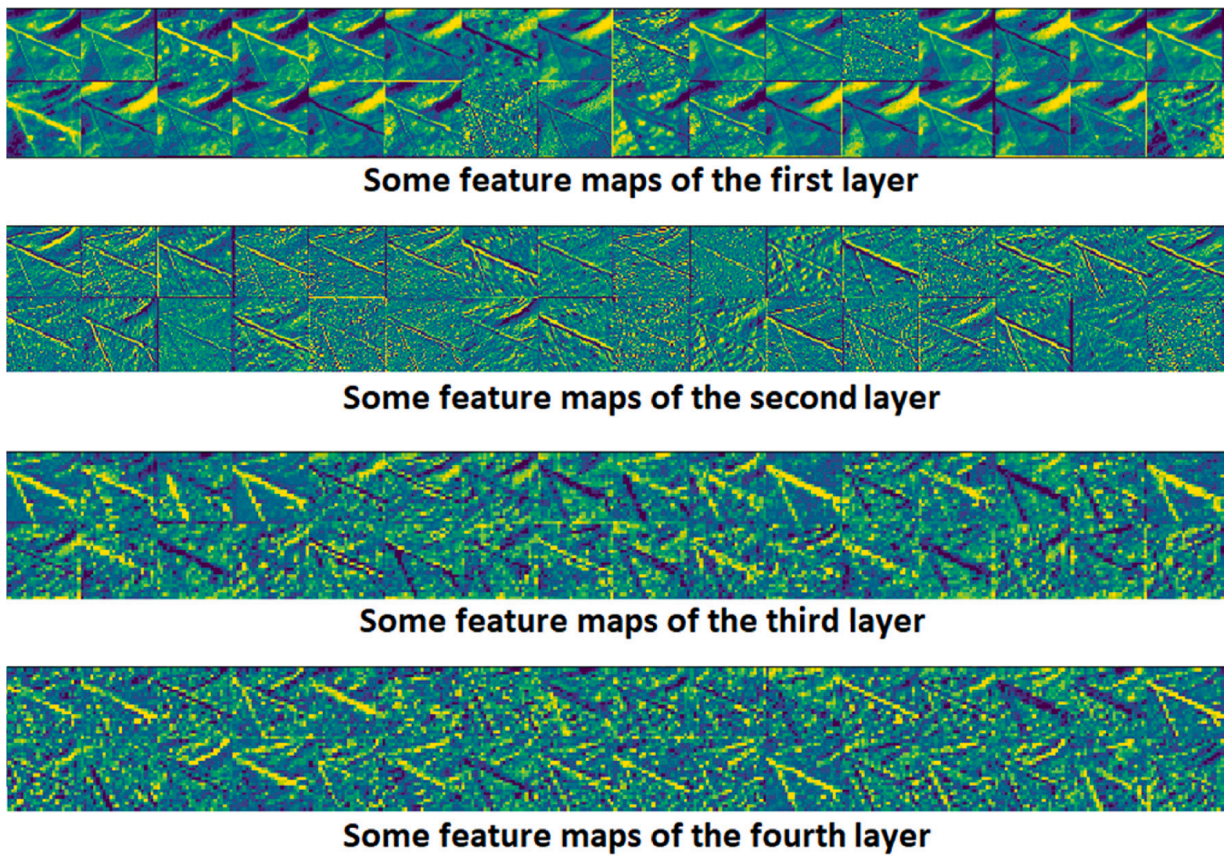

**Fig. 12.** A Number of Feature Maps of the First Four Layers of the P-LeafNET Model Regularized on the Colored Image Patches of MalayaKew (MK). The Third Model Focused only on Venation.

o min_prob_patch: obtaining the minimum probability through the predicted class, or
o min_delta_patch: reaching the minimum probability distance between the predicted class and the second plausible class (probability distance of top-1 from top-2)

If the first condition is met along with at least another condition, the result of the P-LeafNET model is regarded as the final result. Otherwise, the system fails to introduce the definite result and resorts to reporting the most plausible species through the two following approaches:

- vote_rate: initial voting between the classes predicted by the third model on P image patches, or
- vote_merge_rate: general voting on the outputs of three models in case the initial voting result is deficient, and the sufficient minimum probability is obtained from the first and second models.

The termination conditions of the first, second, and third models clearly indicate the effects of using the initial knowledge obtained from the first and second steps on making decisions about announcing the results obtained from the third model. According to the first condition, which must be met, although the first and second models have not correctly introduced the right class as the first choice, their first N and M guesses are still important and effective in the final result. The second and third conditions are for a time when the first and second models fail to meet the minimum certainty conditions, despite guessing the right class. In this case, the result will be accepted less sensitively by confirming this guess through the P-LeafNET model.

## 4. Evaluation of proposed method

This section addresses the performance of the proposed method by conducting tests on the two famous datasets, i.e., MalayaKew (MK) and Flavia, after introducing the training configuration of networks. Then it is compared with other methods. In addition, visualization techniques are employed to analyze different layers of each model and show how much the features of our interest have succeeded in making the proposed method a botanist's method.

### 4.1. Training configurations and settings

This paper used nearly the same training settings for the models employed in the proposed method. These settings are dealt with in detail here. As a result, the model name is referred to only when models differ in training configuration.

- **Cost Function:** The categorical cross-entropy function is used.
- **Optimizer Algorithm:** The Adam (Kingma & Ba, 2014) optimization algorithm is used in this paper.
- **Learning Rate:** The cyclical learning rate (CLR) has been employed to regularize the learning rate (Smith, 2017). In fact, this learning regularization policy increases the basic value of the learning rate in a cyclical mode. In the paper reviewed by (Smith, 2017), the authors showed that CLR could accelerate convergence in the neural network architectures. This policy, with a value of 0.001, is used as the basic learning rate. It is used with the value of 0.006 as the maximum learning rate when the triangular2 mode is employed with the strider of 40.
- **Weight Initialization:** The homogenous Xavier is used in the first and second models for weight initialization (Glorot & Bengio, 2010).
- **Regulator:** In this paper, an L2 regulator is used with a parameter value of 0.001.
- **Batch Size and Epoch:** For each of the three models, the number of epochs is considered 10,000 to store the weights having the highest accuracy on the validation set. The batch sizes of the first, second, and third models are 256, 128, and 512, respectively.

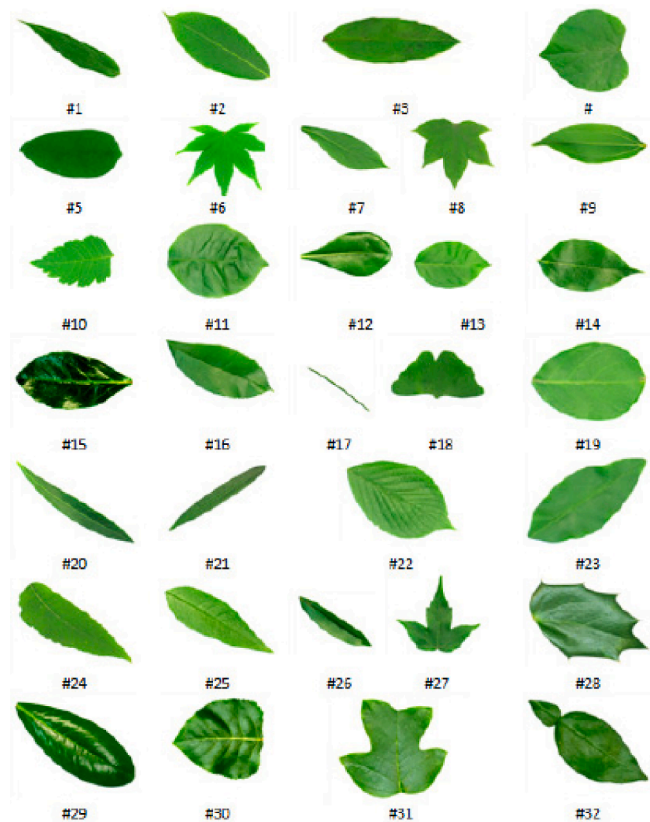

**Fig. 13.** A Sample of Leaf Images Pertaining to Flavia (Wu et al., 2007).

The accuracy of the testing dataset is regarded as the performance evaluation criterion of the proposed method in comparison with other techniques. This criterion is obtained by dividing the number of correct predictions on the testing dataset by the total number of testing datasets. Python is employed to perform the software tasks along with its dedicated deep learning libraries, such as Tensorflow and Keras. Moreover, Nvidia GeForce GTX 1080 8 GB is used in the evaluation system.

### 4.2. Performance analysis of proposed system on MalayaKew

This section first introduces the leaf dataset MalayaKew (MK) (Lee et al., 2017; Lee et al., 2015) and then analyzes the results on it. After that, the results are compared with those of other methods on the same dataset.

#### 4.2.1. MalayaKew (MK) introduction and preparation
The leaf dataset MalayaKew (MK) is collected from Royal Botanic Gardens, Kew, England. This dataset includes the scanned images of leaves in 44 classes. This dataset is very challenging because the leaves of classes of its different species resemble each other greatly. Fig. 8 shows a sample of this leaf dataset.

This dataset includes two subsets, i.e., MK-D1 and MK-D2:

- MK-D1: This dataset includes segmented leaf images sized 256x256. It consists of 2288 and 528 training and testing images, respectively.
- MK-D2: This dataset includes leaf images sized 256x256 pixels. It consists of 34672 and 8800 training and testing images, respectively.

In this dataset, each image has been rotated for 45, 90, 135, 180, 225, 270, and 315 degrees in seven different directions to enhance leaf images.

For training the S-LeafNET model, it is necessary to use binary images, on which the background and entire leaves are black and white, respectively. Therefore, OpenCV and Pillow are used as image processing tools to generate these images from 2288 colored images dedicated for training. For all three networks, all the inputs are divided by 255 because the maximum number of inputs is 255. To ensure learning improvement and generalizability of the designed networks, different data augmentation techniques such as 45-degree rotation, transverse displacement for 0.1 of the entire image, longitudinal displacement for 0.1 of the entire image, horizontal mirror, vertical mirror, and shuffle techniques are employed.

Hence, 2288 images are employed for training the first and second models, whereas 34672 images are utilized for training the P-LeafNET model. For each epoch of 10000 epochs, a combination of data augmentation techniques is randomly applied to images to prevent the overfitting problem in addition to enhancing the generalizability of models. It is impossible to say how many unique images each model experienced during the training process. However, it is clear that all the training datasets are employed along with images generated from data augmentation based on the main images. Furthermore, the entire system input included 528 images of the testing dataset in the testing step. In addition, 528 images of the testing dataset are used separately to evaluate the first and second models, and 8800 image patches are considered in the dataset to test the P-LeafNET model.

**Table 4**
Accuracy Rates of Different Methods on Flavia.

| S. No. | Publications | Method | Accuracy on the entire data (Accuracy on the processed data) |
|---|---|---|---|
| 1 | **Proposed method (Combine)** | **SWP-LeafNET** | **99.67%** |
| 2 | Proposed method (First model) | S-LeafNET | 88.81% (99.76%) |
| 3 | Proposed method (Second model) | W-LeafNET | 95.07% (99.14%) |
| 4 | Proposed method (Third model) | P-LeafNET | 97.72% (100%) |
| 5 | (Sun et al., 2017) | ResNet26 | 99.65% |
| 6 | (Lee et al., 2017) | DeepPlant + MLP | 99.40% |
| 7 | (Saleem et al., 2019) | Shape & Statistical & Vein Features, PCA + KNN | 98.75% |
| 8 | (Barré et al., 2017) | LeafNet CNN | 97.90% |
| 9 | (Naresh & Nagendraswamy, 2016) | Modified LBP (NN) | 97.60% |
| 10 | (Wang et al., 2016) | PCNN + SVM | 96.97% |
| 11 | (Goyal & Kumar, 2018) | 12 features (7 shape, 5 vein). PCA + SVM | 88.79% |

*4.2.2. Evaluation of proposed method on MalayaKew (MK)*

After finishing training each of the three models used in the proposed method implemented on MalayaKew (MK), it is necessary to integrate the three models to develop the proposed system. The second column of Table 2 shows the parameters introduced in the previous section for the termination of each step, and the previous knowledge transferred to the next step for MalayaKew (MK).

Separate tests are conducted on each of the models introduced by the proposed method to obtain 96.59%, 97.35%, and 95.35% of accuracy for the first, second, and third models, respectively. According to Table 3, these models showed no defendable performance in comparison with the most-recent algorithms in terms of accuracy. However, it is worth mentioning to note that we have never sought to use a single model separately. In other words, our goal is to maximize the accuracy of each model on the data that the model has enough confidence in its predicted class and has decided to report it as the final result, not on all data where it includes data that is due to insufficient certainty about its class is transferred to next models. Given this assumption, the accuracy of each model on the processed data is 99.77%, 100%, and 100%, respectively.

In particular, when the proposed system encounters 528 MalayaKew (MK) datasets, the S-LeafNET model makes the final decision on 441 images (i.e., 83.52% of the total data). The process of analyzing these images ends because the termination conditions of the first step are met. Out of 441 images, only one image is mispredicted. This false predicted image pertains to class 2, and the system attributed it wrongly to class 27. Fig. 9 shows this sample with an image of the true class and an image of the false class. Obviously, similarity to class 27 is much stronger than similarity to the correct class 2; therefore, the system's decision matches the decision made by observing both classes.

Next, the 87 images on which the S-LeafNET model is unable to make firm decisions are given to the W-LeafNET model, which decided on 78 of them (i.e., 14.77% of the whole data). Fortunately, the decisions of the W-LeafNET model on these 78 images are totally right with no errors. Finally, the nine remaining images are given to the P-LeafNET model (i.e., 1.71% of the entire data). After the automated extraction of image patches from each of these nine images, they are predicted. In this step, the prediction is performed with no errors. Eventually, it is fair to say that the proposed model allocated 527 of 528 test images of MalayaKew (MK) correctly to the correct class.

In other words, after integrating these three models and developing the proposed system, accuracy reached a perfect rate of 99.81%. According to Table 3, the proposed method outperformed all the other methods evaluated on MalayaKew (MK). In addition, since these models are characterized by shorter depth and fewer parameters than the other deep learning-based models, they can result in higher speeds and reduce the memory space required for storage and processing.

Specifically, compared with conventional methods (9–13), the proposed method achieved much higher accuracy rates. In addition, it requires no hand-crafted feature extraction, which has specific challenges and depends greatly on datasets. Compared with deep learning-based methods (5–7), one might wonder if the superiority of the proposed algorithm lies only in terms of accuracy improvement.

To answer this question, the proposed method is much faster than other methods, benefiting from the pre-trained AlexNet (6 & 7) or the simultaneous integration of a multi-input network (5) due to the use of shallower networks with fewer parameters and also separate use of each model. It further enhances accuracy. The most important superiority of the proposed system is that its guesses can be used as a botanist's opinions when unknown species or genetic changes are observed. This is because we have accurately simulated the behavior of a botanist and have succeeded in providing an interpretable model that is reliable.

Another advantage of the suggested approach is distributablity. In other words, as we mentioned before, the first model could be implemented on a user's mobile phone, whereas the second model is on a user's computer, and the third model is installed on a server. Now, most images can successfully be processed on the user's mobile phone, and there is no need to communicate with the next models. At the same time, the second and third models are used in the case of more complicated images. This very simple distributablity of the suggested algorithm is absent in all the compared deep learning-based algorithms. The only flaw of the proposed method is the system complexity in training and regularizing termination parameters, and transferring previous knowledge to the next step. However, all of these problems should be dealt with by developers, and users are unaware of them while using and testing the system, which can be used easily.

It should also be mentioned that the parameters of Table 2 are obtained empirically and that it is decided to strike a balance between the minimum and maximum speeds. In other words, it is decided to evaluate most of the images definitively in the first step by resorting to the second and third models less often so that the algorithm could run at the maximum power. At the same time, it is meant to reach the maximum accuracy possible. Obviously, if more images are entered into the second and third models for evaluation, the model reliability will increase; however, the algorithm execution speed will decrease. Instead, if the first model is used as much as possible, the proposed system will have a lower reliability rate while the algorithm execution speed reaches the maximum rate.

**Table 5**
Results of Five Tests Conducted on 304 Testing Data Selected Randomly from Flavia.

| Test number | Test 1 | Test 2 | Test 3 | Test 4 | Test 5 | Average |
|---|---|---|---|---|---|---|
| The Number of Samples Evaluated by the First Model (the number of samples evaluated correctly by the S-LeafNET model) | 252 (252) | 251 (250) | 252 (252) | 250 (250) | 252 (250) | 251.4 (250.8) |
| The Number of Samples Evaluated by the Second Model (the number of samples evaluated correctly by the W-LeafNET model) | 45 (45) | 44 (43) | 46 (45) | 48 (48) | 50 (50) | 46.6 (46.2) |
| The Number of Samples Evaluated by the Third Model (the number of samples evaluated correctly by the P-LeafNET model) | 7 (7) | 9 (9) | 6 (6) | 6 (6) | 2 (2) | 6 (6) |
| **General Accuracy of Proposed Method** | **100%** | **99.34%** | **99.67%** | **100%** | **99.34%** | **99.67%** |

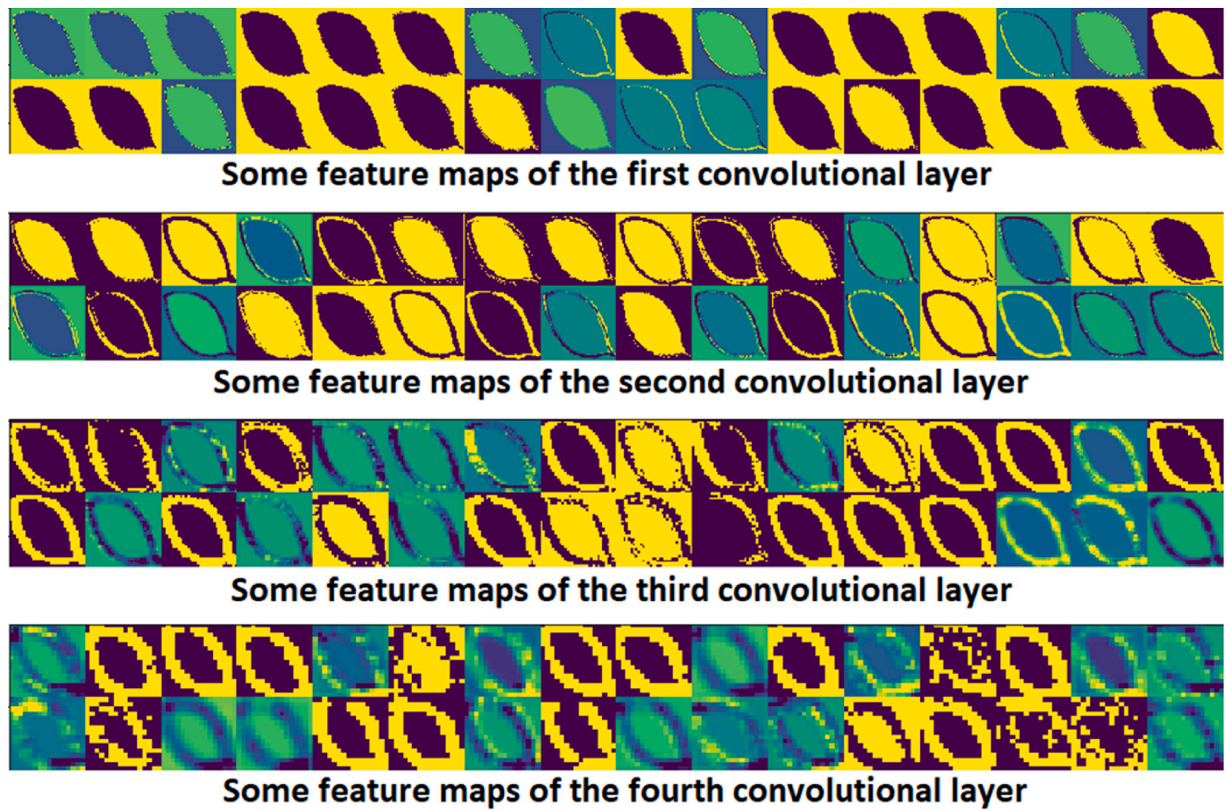

**Fig. 14.** A Number of Feature Maps of the First Four Convolutional Layers of the S-LeafNET Model Trained on the Black-and-White Images Obtained from Flavia. The First Model Focused only on the General Shape of the Leaf.

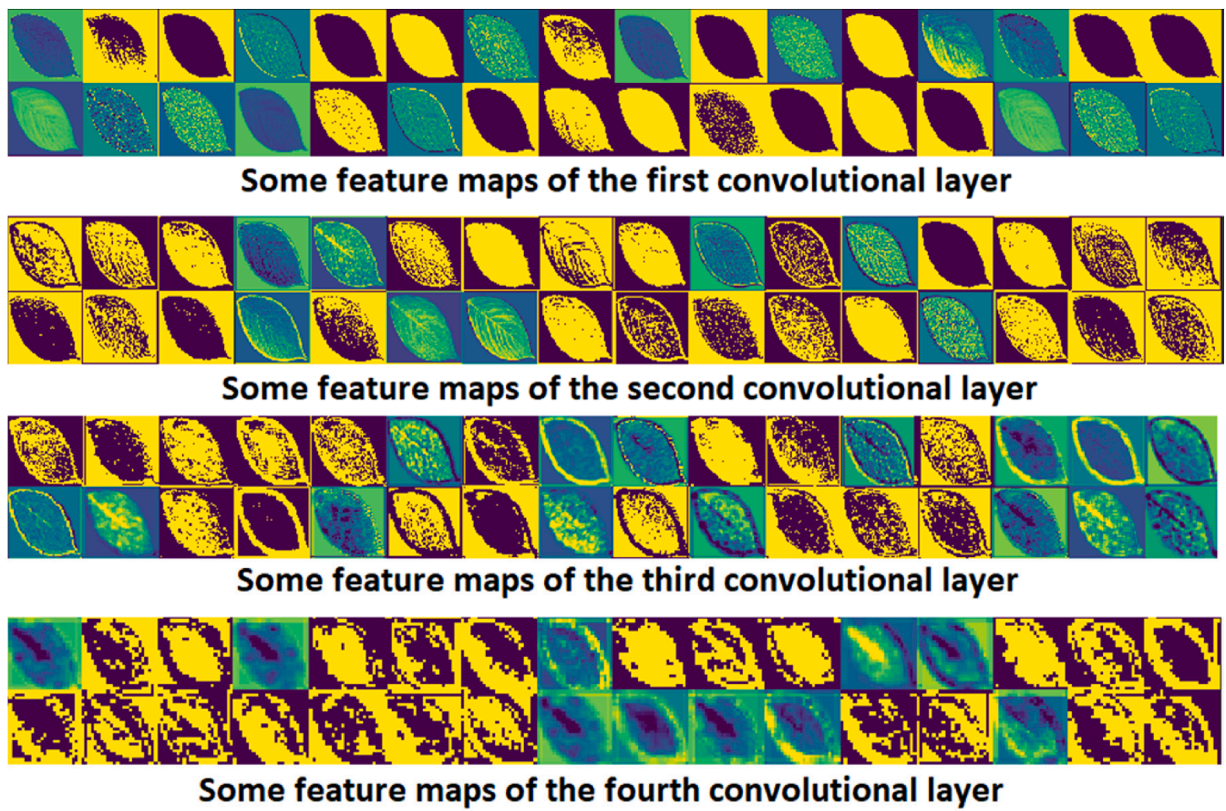

**Fig. 15.** A Number of Feature Maps of the First Four Convolutional Layers of the W-LeafNET Model Trained on the Colored Images Obtained from Flavia. The Second Model Took an Extensive Look at the Leaf. It Paid More Attention to Color and Venation, Although the General Shape of the Leaf is Much More Important to it.

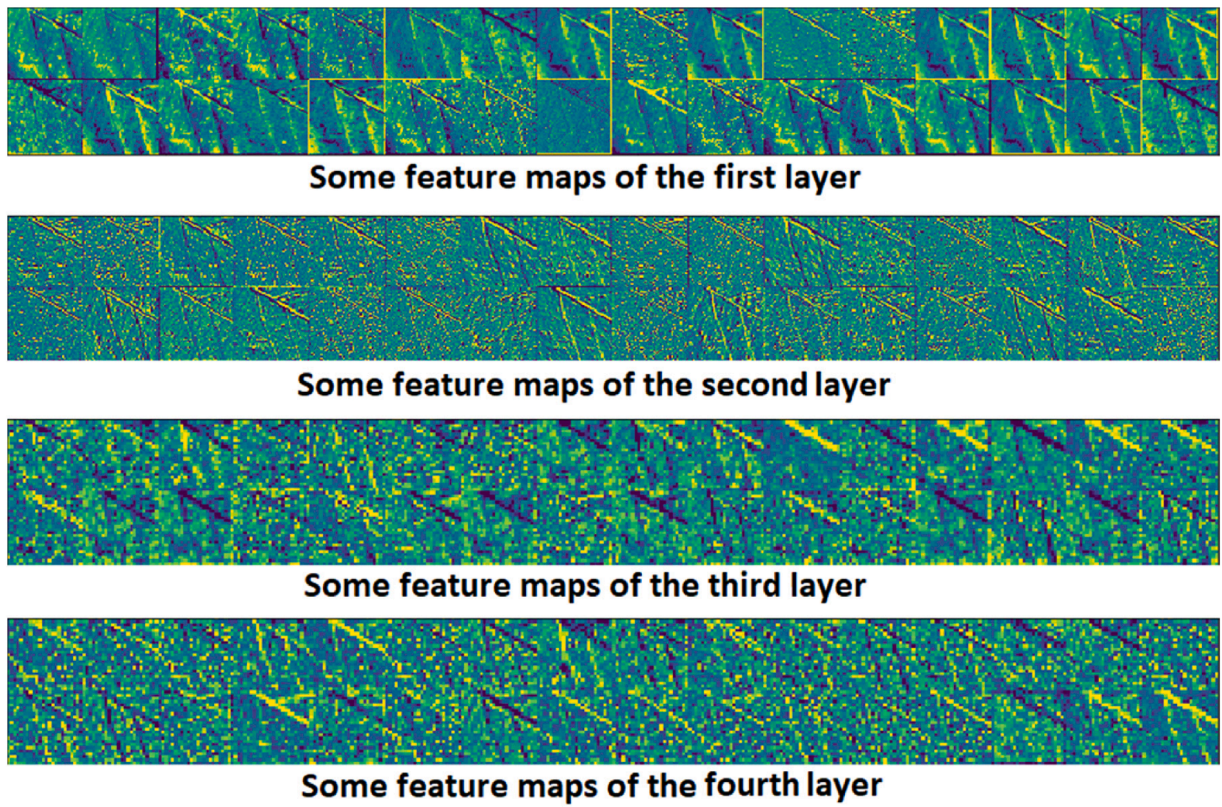

**Fig. 16.** A Number of Feature Maps of the First Four Layers of the P-LeafNET Model Regularized on the Image Patches Obtained from Flavia. The Third Model Focused only on Venation.

The higher reliability rate of the three models is defendable because each of these three models experienced a different view of data. Thus, integrating these distinctively trained features can significantly help distinguish the right class more reliably. For this purpose, all layers of the three models are visualized to show what leaf features each model focused on and if it is possible to model a botanist's behavior. Fig. 10, Fig. 11, and Fig. 12 show the images of different layers from the first, second, and third models, respectively. Accordingly, the first model focused only on the general shape of the leaf. However, the second model took a general look at the leaf. It also analyzed the color and venation, although the general shape of the leaf is more important. Finally, the P-LeafNET model focused only on venation. Thus, botanist's behavior is successfully modeled. In other words, now we are sure it makes decisions like a botanist. As a result, despite the use of deep learning-based models, one can ensure that the proposed system is reliable and there is no concern about the lack of interpretability of the proposed method.

### 4.3. Performance evaluation of proposed method on Flavia

After introducing Flavia (Wu et al., 2007), this section analyzes the results of the proposed method and compares them with those of other methods on this dataset. This evaluation aims to show the capability of the proposed system in the face of different datasets and its absence from having a bias to the database. With this experiment, the generalizability of the proposed method can be confirmed.

#### 4.3.1. Introducing and preparing Flavia

Flavia is another well-known leaf identification dataset introduced by (Wu et al., 2007). It consists of 1907 leaf images of 32 different species. The images are sized 1600x1200 pixels. Fig. 13 shows one of them.

To evaluate the proposed method on this dataset, binary images are required for training the S-LeafNET model. The background of these images should be black, whereas the entire leaf should be white. Image processing tools such as OpenCV and Pillow are employed to generate these images from 1907 dedicated colorful training images. For training the P-LeafNET model, image patches are obtained from the main images. As a result, 801 image patches are automatically generated for each class. The leaves accounted for at least 98% of each image. Therefore, there are 25,632 image patches generated for training the P-LeafNET model. For each of the three networks, all the inputs are divided by 255 because the maximum numerical value of inputs is 255. Rotation up to 45 degrees, traverse displacement up to 0.1 of the entire image, longitudinal displacement up to 0.1 of the entire image, horizontal mirror, vertical mirror, and shuffle techniques are performed to ensure learning improvement and generalizability of the designed networks.

Since Flavia lacks separate sections for training and testing, 16.7% of this dataset (nearly ten samples of each class) is randomly separated to develop a testing dataset in order to compare the results fairly with those of other methods. Hence, 1603 images are employed for training the first and second models, and 21376 image patches are utilized for training the P-LeafNET model. For each epoch of 10000 epochs, a random combination of data augmentation techniques is applied to the images to enhance the generalizability of models and prevent the overfitting problem. As a result, it is impossible to say how many unique images each model experienced in the training process. However, it is obvious that all the training datasets are used along with the images generated through data augmentation based on the main images. In addition, the entire system input included 304 testing dataset images generated randomly. Furthermore, 304 testing dataset images are developed separately to test the first and second models. However, 4256 image patches are generated randomly from a total number of 25632 images to test the P-LeafNET model. The process is repeated five times despite the random separation of the training dataset from the testing dataset. The reported results are obtained from the mean of these tests.

*4.3.2. Evaluation of proposed method on Flavia*

After finishing training each of the three models on Flavia, it is time to integrate them to achieve the proposed method. The third column of Table 2 shows the parameters introduced in the previous section for the termination of each step, and the previous knowledge transferred to the next step for Flavia.

Separate tests are conducted on each of the three models in the proposed method to show that the first, second, and third models achieve 88.81%, 95.07%, and 97.72% accuracy rates, respectively. According to Table 4, these models showed no defendable performance in comparison with the latest algorithms in terms of accuracy. However, keep in mind that we have never used a single model separately.

More fully, the performance is evaluated in the mean accuracy obtained from each model after conducting five tests. Table 5 shows each of the five tests for a more detailed analysis of results. When the proposed method encounters 304 testing data obtained randomly from Flavia, it has a relative equilibrium on every five tests. It meets the termination conditions of the first model with nearly 83% of data and decides on them. Regarding 15% of data, the second model is sufficient, and there is no need to continue the algorithm. Therefore, only 2% of the data will continue the algorithm to the end. On average, the first model managed to make accurate decisions on data, meeting the termination conditions of the first step with a success rate of 99.76%. The success rates of the second and third models are 99.14% and 100%, respectively. These results point more fairly to the intelligence of the proposed system, where our system is well able to identify when it needs the next models. Finally, by integrating these three models, the proposed model has managed to correctly identify 99.67% of testing data from Flavia on average. As shown in Table 4, it outperforms all the other methods evaluated on this dataset.

As discussed previously, these models have fewer parameters than the other deep learning-based models. As a result, it is applicable for limited computational and storage recourses like portable devices. Moreover, in comparison with traditional methods (7, 9, 10, and 11), the proposed method does not need hand-crafted feature extracting, and its performance does not rely on the dataset. In comparison with deep learning-based methods (5, 6, and 8), the proposed method performed much faster than other pre-trained models such as ResNet26 (5), AlexNet (6), and ResNet50 (8) because of using shallower networks with fewer parameters and also using all three models distinctively. It also enhanced accuracy compared to all other approaches. As discussed in the analysis of the previous dataset, another advantage of the proposed method lies in the successful modeling of a botanist's behavior and its distributablity. Consequently, the proposed system is not only accurate but also reliable and trustworthy for botanists.

In the end, all layers of the three models are visualized again on this dataset to show what leaf features each model focused on. As are shown in Figs. 14–16, the S-LeafNET model focused only on the general shape of the leaf. However, the W-LeafNET model took a more extensive look at the leaf. It also paid attention to color and venation, even though the general shape of the leaf is much more important to it. Finally, the P-LeafNET model focused only on venation. These steps are the same as a botanist in a botanical lab does, except that now not only is there no need to cut the leaf and carry it to the lab, but it can be done quickly onsite. In addition, by performing similar measures and visualizations on the Flavia dataset, the generalizability and performance stability of the proposed method can give botanists more confidence in the proposed method.

## 5. Conclusion

This paper proposed a method with a maximum behavioral resemblance with a botanist's behavior. Three deep learning-based models (S-LeafNET, W-LeafNET, and P-LeafNET) are employed to develop the proposed method. The first and second models are designed from scratch, and the third model employed the pre-trained MobileNetV2 model. The tests are conducted on the two well-known datasets of leaf identification, i.e., MalayaKew (MK) and Flavia. According to the results, the proposed method obtained 99.81% and 99.67% of accuracy and outperformed all the other methods. Compared to conventional methods, the suggested approach required no hand-crafted feature extraction, having specific complexities and depending greatly on datasets. In comparison with other deep learning-based techniques, the proposed method enhanced accuracy. It also acted much faster than other methods because of using shallower networks, fewer parameters, and further using the three models repeatedly.

The most important superiority of the proposed system is that its guesses can be used as a botanist's theories when unknown species or genetic changes are observed because the method is developed to model a botanist's behavior. To prove this, visualization techniques are employed to analyze different layers of each model and show that the expected features successfully model a botanist's behavior. Another advantage is the distributablity of the proposed method. Compared to other methods, the only flaw is the system complexity in training and regulating termination parameters and transferring previous knowledge to the next step. Nonetheless, it is fair to say that all of these problems will be experienced by developers, and users will be unaware of any difficulties in using and testing the system. Thus, users can easily use the system.

In the future, it is recommended to obtain the optimal and semi-optimal values of termination parameters through heuristic and meta-heuristic algorithms by defining an appropriate cost function to increase accuracy and pass fewer steps in a bid to perfect the proposed method. In addition, it is possible to identify other components of plants such as roots, flowers, and stems simultaneously to develop a hybrid system to enhance accuracy and efficiency in plant identification. It is also necessary to use complicated datasets obtained from real environmental conditions and establish a system to identify images successfully.

*CRediT authorship contribution statement*

**Ali Beikmohammadi:** Conceptualization, Methodology, Software, Validation, Investigation, Writing – original draft, Writing – review & editing, Visualization. **Karim Faez:** Resources, Data curation, Supervision. **Ali Motallebi:** Writing – original draft.

**Declaration of Competing Interest**

The authors declare that they have no known competing financial interests or personal relationships that could have appeared to influence the work reported in this paper.

## References

Angelov, P., & Sperduti, A. (2016). Challenges in deep learning. In: *ESANN 2016 - 24th European Symposium on Artificial Neural Networks*, 489-496.

Barré, P., Stöver, B. C., Müller, K. F., & Steinhage, V. (2017). LeafNet: A computer vision system for automatic plant species identification. *Ecological Informatics, 40*, 50–56.

Beikmohammadi, A., & Faez, K. (2018, December). Leaf classification for plant recognition with deep transfer learning. In *2018 4th Iranian Conference on Signal Processing and Intelligent Systems (ICSPIS)*, 21-26. IEEE. doi: 10.1109/icspis.2018.8700547.

Bodhwani, V., Acharjya, D. P., & Bodhwani, U. (2019). Deep residual networks for plant identification. *Procedia Computer Science, 152*, 186–194.

Charters, J., Wang, Z., Chi, Z., Tsoi, A. C., & Feng, D. D. (2014, July). EAGLE: A novel descriptor for identifying plant species using leaf lamina vascular features. In *2014 IEEE international conference on multimedia and expo workshops (ICMEW)*, 1-6. IEEE.

Chollet, F. (2017). Xception: Deep learning with depthwise separable convolutions. In *Proceedings of the IEEE conference on computer vision and pattern recognition* (pp. 1251–1258).

Clarke, J., Barman, S., Remagnino, P., Bailey, K., Kirkup, D., Mayo, S., & Wilkin, P. (2006, November). Venation pattern analysis of leaf images. In *International Symposium on Visual Computing*, 427-436. Springer.

Cope, J. S., Corney, D., Clark, J. Y., Remagnino, P., & Wilkin, P. (2012). Plant species identification using digital morphometrics: A review. *Expert Systems with Applications, 39*(8), 7562–7573.

Cope, J. S., Remagnino, P., Barman, S., & Wilkin, P. (2010, November). Plant texture classification using gabor co-occurrences. In *International Symposium on Visual Computing*, 669-677. Springer.

Du, C., & Gao, S. (2017). Image segmentation-based multi-focus image fusion through multi-scale convolutional neural network. *IEEE Access, 5*, 15750–15761.

Ferentinos, K. P. (2018). Deep learning models for plant disease detection and diagnosis. *Computers and Electronics in Agriculture, 145*, 311–318.

Glorot, X., & Bengio, Y. (2010, March). Understanding the difficulty of training deep feedforward neural networks. In *Proceedings of the thirteenth international conference on artificial intelligence and statistics*, 249-256. JMLR Workshop and Conference Proceedings.

Goyal, N., & Kumar, N. (2018, September). Plant species identification using leaf image retrieval: A study. In *2018 International Conference on Computing, Power and Communication Technologies (GUCON)*, 405-411. IEEE.

Grinblat, G. L., Uzal, L. C., Larese, M. G., & Granitto, P. M. (2016). Deep learning for plant identification using vein morphological patterns. *Computers and Electronics in Agriculture, 127*, 418–424.

Hall, D., McCool, C., Dayoub, F., Sunderhauf, N., & Upcroft, B. (2015, January). Evaluation of features for leaf classification in challenging conditions. In *2015 IEEE Winter Conference on Applications of Computer Vision*, 797-804. IEEE.

Hedjazi, M. A., Kourbane, I., & Genc, Y. (2017, May). On identifying leaves: A comparison of CNN with classical ML methods. In *2017 25th Signal Processing and Communications Applications Conference (SIU)*, 1-4. IEEE.

Hu, J., Chen, Z., Yang, M., Zhang, R., & Cui, Y. (2018). A multiscale fusion convolutional neural network for plant leaf recognition. *IEEE Signal Processing Letters, 25*(6), 853–857.

Ioffe, S., & Szegedy, C. (2015, June). Batch normalization: Accelerating deep network training by reducing internal covariate shift. In *International conference on machine learning*, 448-456. PMLR.

Kalyoncu, C., & Toygar, Ö. (2015). Geometric leaf classification. *Computer Vision and Image Understanding, 133*, 102–109.

Kamilaris, A., & Prenafeta-Boldú, F. X. (2018). Deep learning in agriculture: A survey. *Computers and electronics in agriculture, 147*, 70–90.

Kingma, D. P., & Ba, J. (2014). Adam: A method for stochastic optimization. *arXiv preprint arXiv:1412.6980*.

Kumar, N., Belhumeur, P. N., Biswas, A., Jacobs, D. W., Kress, W. J., Lopez, I. C., & Soares, J. V. (2012, October). Leafsnap: A computer vision system for automatic plant species identification. In *European conference on computer vision*, 502-516. Springer.

Larese, M. G., Namías, R., Craviotto, R. M., Arango, M. R., Gallo, C., & Granitto, P. M. (2014). Automatic classification of legumes using leaf vein image features. *Pattern Recognition, 47*(1), 158–168.

Lee, S. H., Chan, C. S., Mayo, S. J., & Remagnino, P. (2017). How deep learning extracts and learns leaf features for plant classification. *Pattern Recognition, 71*, 1–13.

Lee, S. H., Chan, C. S., Wilkin, P., & Remagnino, P. (2015, September). Deep-plant: Plant identification with convolutional neural networks. In *2015 IEEE international conference on image processing (ICIP)*, 452-456. IEEE.

Liu, Z., Zhu, L., Zhang, X. P., Zhou, X., Shang, L., Huang, Z. K., & Gan, Y. (2015, August). Hybrid deep learning for plant leaves classification. In *International Conference on Intelligent Computing*, 115-123. Springer.

Mouine, S., Yahiaoui, I., & Verroust-Blondet, A. (2012, June). Advanced shape context for plant species identification using leaf image retrieval. In *Proceedings of the 2nd ACM international conference on multimedia retrieval*, 1-8.

Naresh, Y. G., & Nagendraswamy, H. S. (2016). Classification of medicinal plants: An approach using modified LBP with symbolic representation. *Neurocomputing, 173*, 1789–1797.

Neto, J. C., Meyer, G. E., Jones, D. D., & Samal, A. K. (2006). Plant species identification using Elliptic Fourier leaf shape analysis. *Computers and electronics in agriculture, 50*(2), 121–134.

Rasti, R., Rabbani, H., Mehridehnavi, A., & Hajizadeh, F. (2017). Macular OCT classification using a multi-scale convolutional neural network ensemble. *IEEE Transactions on Medical Imaging, 37*(4), 1024–1034.

Saleem, G., Akhtar, M., Ahmed, N., & Qureshi, W. S. (2019). Automated analysis of visual leaf shape features for plant classification. *Computers and Electronics in Agriculture, 157*, 270–280.

Sandler, M., Howard, A., Zhu, M., Zhmoginov, A., & Chen, L. C. (2018). Mobilenetv 2: Inverted residuals and linear bottlenecks. In *Proceedings of the IEEE conference on computer vision and pattern recognition* (pp. 4510–4520).

Shelhamer, E., Long, J., & Darrell, T. (2016). Fully convolutional networks for semantic segmentation. *IEEE Transactions on Pattern Analysis and Machine Intelligence, 39*(4), 640–651.

Simonyan, K., & Zisserman, A. (2014). Very deep convolutional networks for large-scale image recognition. *arXiv preprint arXiv:1409.1556*.

Smith, L. N. (2017, March). Cyclical learning rates for training neural networks. In *2017 IEEE winter conference on applications of computer vision (WACV)*, 464-472. IEEE.

Srivastava, N., Hinton, G., Krizhevsky, A., Sutskever, I., & Salakhutdinov, R. (2014). Dropout: A simple way to prevent neural networks from overfitting. *The Journal of Machine Learning Research, 15*(1), 1929–1958.

Su, Y. C., Chiu, T. H., Yeh, C. Y., Huang, H. F., & Hsu, W. H. (2014). Transfer learning for video recognition with scarce training data for deep convolutional neural network. *arXiv preprint arXiv:1409.4127*.

Sun, Y., Liu, Y., Wang, G., & Zhang, H. (2017). Deep learning for plant identification in natural environment. *Computational intelligence and neuroscience, 2017*.

Szegedy, C., Ioffe, S., Vanhoucke, V., & Alemi, A. A. (2017). February). Inception-v4, inception-resnet and the impact of residual connections on learning. *Thirty-first AAAI conference on artificial intelligence*.

Szegedy, C., Vanhoucke, V., Ioffe, S., Shlens, J., & Wojna, Z. (2016). Rethinking the inception architecture for computer vision. In *Proceedings of the IEEE conference on computer vision and pattern recognition* (pp. 2818–2826).

Tang, Z., Su, Y., Er, M. J., Qi, F., Zhang, L., & Zhou, J. (2015). A local binary pattern based texture descriptors for classification of tea leaves. *Neurocomputing, 168*, 1011–1023.

Wang, Z., Sun, X., Zhang, Y., Ying, Z., & Ma, Y. (2016). Leaf recognition based on PCNN. *Neural Computing and Applications, 27*(4), 899–908.

Wu, S. G., Bao, F. S., Xu, E. Y., Wang, Y. X., Chang, Y. F., & Xiang, Q. L. (2007, December). A leaf recognition algorithm for plant classification using probabilistic neural network. In *2007 IEEE international symposium on signal processing and information technology*, 11-16. IEEE.

Xiao, X. Y., Hu, R., Zhang, S. W., & Wang, X. F. (2010, August). HOG-based approach for leaf classification. In *International Conference on Intelligent Computing*, 149-155. Springer.

Yang, J., Yu, K., Gong, Y., & Huang, T. (2009, June). Linear spatial pyramid matching using sparse coding for image classification. In *2009 IEEE Conference on computer vision and pattern recognition*, 1794-1801. IEEE.